\documentclass[
]{ceurart}

\usepackage{listings}
\usepackage[most]{tcolorbox}
\begin{document}

\copyrightyear{2025}
\copyrightclause{Copyright for this paper by its authors.
  Use permitted under Creative Commons License Attribution 4.0
  International (CC BY 4.0).}

\conference{CLEF 2025 Working Notes, 9 -- 12 September 2025, Madrid, Spain}
\title{ClaimIQ at CheckThat! 2025: Comparing Prompted and Fine-Tuned Language Models for Verifying Numerical Claims}

\title[mode=sub]{Notebook for the CheckThat! Lab at CLEF 2025}


\author[1]{Anirban Saha Anik}[%
orcid=0000-0002-7824-3702,
email=AnirbanSahaAnik@my.unt.edu,
]
\cormark[1]
\fnmark[1]

\author[2]{Md Fahimul Kabir Chowdhury}[%
orcid=0009-0003-8034-5235,
email=MdFahimulKabirChowdhury@my.unt.edu,
]
\fnmark[1]

\author[2]{Andrew Wyckoff}[%
email=AndrewWyckoff@my.unt.edu,
]

\author[2]{Sagnik Ray Choudhury}[%
email=Sagnik.Choudhury@unt.edu,
]
\address[1]{Department of Data Science, University of North Texas, Denton, TX, USA}
\address[2]{Department of Computer Science and Engineering, University of North Texas, Denton, TX, USA}

\cortext[1]{Corresponding author.}
\fntext[1]{These authors contributed equally.}

\begin{abstract}
This paper presents our system for Task 3 of the CLEF 2025 CheckThat! Lab, which focuses on verifying numerical and temporal claims using retrieved evidence. We explore two complementary approaches: zero-shot prompting with instruction-tuned large language models (LLMs) and supervised fine-tuning using parameter-efficient LoRA. To enhance evidence quality, we investigate several selection strategies, including full-document input and top-$k$ sentence filtering using BM25 and MiniLM. Our best-performing model \texttt{LLaMA} fine-tuned with LoRA achieves strong performance on the English validation set. However, a notable drop in the test set highlights a generalization challenge. These findings underscore the importance of evidence granularity and model adaptation for robust numerical fact verification.

\end{abstract}


\maketitle

\section{Introduction}

As misinformation continues to spread across digital platforms, the ability to automatically verify factual claims has become increasingly important ~\cite{figueira2017current}. Among the most challenging forms of misinformation are those involving numerical or temporal elements, claims that reference statistics, quantities, dates, or trends ~\cite{meel2020fake}. These claims are often persuasive and deceptively simple; yet, verifying them requires not just factual knowledge but also precise reasoning over quantitative details.

To support the verification of numerical misinformation, Viswanathan et al.~\cite{venktesh2024quantemp} proposed the QuanTemp dataset. This benchmark targets real-world quantitative and temporal claims, including multilingual evidence retrieved from fact-checking sources. It serves as the foundation for CLEF 2025 Task 3 \cite{clef-checkthat:2025:task3}. Task 3 of the CLEF 2025 CheckThat! Lab \cite{clef2025-workingnotes} focuses on verifying such claims by classifying them as \texttt{True}, \texttt{False}, or \texttt{Conflicting} based on a small set of retrieved evidence. This task is especially challenging because evidence is frequently noisy, partially relevant, or even contradictory, and claims may rely on implicit or contextualized numerical reasoning.

Recent advancements in large language models (LLMs) have shown promising capabilities in understanding and generating human-like text \cite{qian2023harnessing, yu2023generate, wang2022towards}. However, their effectiveness in structured fact verification, especially when reasoning over multiple retrieved evidence passages, remains an open research problem. Additionally, aligning LLM outputs with factual correctness while managing computational efficiency is a key consideration.

In this work, we explore two complementary strategies for numerical claim verification: zero-shot prompting with instruction-tuned LLMs and supervised fine-tuning using parameter-efficient methods (LoRA). We also experiment with various evidence selection techniques, including full-document input and top-$k$ sentence retrieval via BM25 and MiniLM, to assess the impact of evidence granularity on model performance.

Our approach aims to evaluate the balance between generalization and supervision, and to investigate how LLMs can be adapted for precise, scalable, and reliable numerical fact-checking.

\section{Related Work}

Recent years have seen growing interest in fact verification systems that integrate natural language processing, information retrieval, and reasoning ~\cite{lazarski2021using, hong2025dynamic}. A prominent line of work in this space is retrieval-augmented generation (RAG), which combines document retrieval with large language models (LLMs) to produce contextually grounded and factually accurate outputs ~\cite{huang2024survey, anik2025multi}. Yue et al. ~\cite{yue2024evidence} introduced RARG, a retrieval-augmented RAG framework that incorporates scientific literature to generate polite, evidence-based counter-responses. Their use of reinforcement learning with document-level supervision demonstrated the benefits of aligning generation with factual evidence. Expanding on this, RAFTS~\cite{yue2024retrieval} introduced a contrastive fact verification pipeline that generates both supporting and refuting responses from retrieved passages. RAFTS emphasized interpretability and achieved strong results using parameter-efficient models.

Systems such as FactGenius ~\cite{gautam2024factgenius} improve zero-shot prompt-based fact-checking abilities of LLMs by integrating them with external knowledge bases (DBPedia) and similarity measures (fuzzy text matching).
ClaimMatch ~\cite{pisarevskaya2025zero} leverages LLMs in both zero-shot and few-shot settings (e.g., GPT-3.5-turbo, Gemini, LLaMA) for claim matching (CM), utilizing natural language inference and paraphrase detection. Tang et al. ~\cite{tang2024minicheck} developed MiniCheck, a sentence-level verifier that approaches GPT-4 performance using synthetic training data and smaller models. Their work shows that compact models can perform competitively when fine-tuned appropriately.

Several researchers have employed Full-Context Retrieval and Verification frameworks to perform LLM-based claim extraction in conjunction with Retrieval-Augmented Generation (RAG). RAG enhances the detection process by constructing a comprehensive context for fact-checking~\cite{bai2024large,laban2024summary,russo2024face}. 

Our approach builds on these insights by combining sentence-level retrieval (BM25 and MiniLM), fine-tuned generation with LLaMA, and multilingual claim-evidence alignment. Unlike decomposition-heavy pipelines, we show that strong performance can be achieved with simpler architectures and focused supervision.

\section{Task Description}

\subsection{Task Overview}

We participate in \textbf{Task 3: Fact-Checking Numerical Claims} as part of the CLEF 2025 CheckThat! Lab \cite{clef-checkthat:2025-lncs}. This task aims to verify the factual correctness of claims that include numerical quantities or temporal expressions. Such claims require not only linguistic understanding but also the ability to interpret quantities, dates, and time-based facts in context.

Participants are provided with a set of claims and corresponding evidence passages retrieved using top-100 BM25 ranking. The goal is to classify each claim into one of three labels:

\begin{itemize}
    \item \textbf{True} - the claim is fully supported by the evidence;
    \item \textbf{False} - the claim is clearly refuted by the evidence;
    \item \textbf{Conflicting} - the evidence is ambiguous, partially supportive, or contradictory.
\end{itemize}

The task challenges systems to handle ambiguous evidence, resolve conflicting numbers or dates, and reason over concise or incomplete textual data. Participants are allowed to apply re-ranking, retrieval filtering, and generation techniques to improve verification performance.

\subsection{Dataset Summary}

For Task 3, we use a dataset sourced from fact-checking reports gathered via the Google Fact Check Explorer API. We filter claims to include only those with numerical or temporal expressions. Each claim comes with a ranked set of evidence documents, retrieved using BM25 and claim decomposition.*

Though the dataset supports multiple languages, we limit our experiments to the English portion.

\begin{table}[h]
\centering
\small 
\caption{CLEF 2025 Task 3 Dataset Overview (English Subset)}
\begin{tabular}{@{}ll@{}}
\toprule
\textbf{Attribute}       & \textbf{Value}                          \\
\midrule
Language                & English                                 \\
Number of Claims        & 15,514                                  \\
Evidence per Claim      & Top-100 BM25-ranked documents           \\
Labels                  & True, False, Conflicting                \\
Task Format             & 3-class classification                  \\
\bottomrule
\end{tabular}
\end{table}

\section{Methodology}
\subsection{Problem Formulation}

The goal of this task is to automatically verify the factual correctness of claims that contain numerical or temporal expressions. Each instance in the dataset consists of a claim $C$ and a corresponding evidence set $E = \{e_1, e_2, ..., e_k\}$, where each $e_i$ is a sentence or a document retrieved from a fact-checking corpus. The task is to classify the claim into one of three categories: \texttt{True}, \texttt{False}, or \texttt{Conflicting}.

We treat this as a three-way classification problem, where the model learns a function:
\[
f(C, E) \rightarrow y \in \{\texttt{True}, \texttt{False}, \texttt{Conflicting}\}
\]
Here, $f$ can be instantiated as either a generative language model prompted in zero-shot fashion, or a fine-tuned discriminative classifier.

The evidence $E$ is varied across different experimental configurations. In some cases, $E$ includes the full document retrieved via BM25, while in others, it consists of a ranked subset of top-$k$ relevant sentences, or a summary generated by a large language model. This flexible formulation allows us to investigate the effect of evidence selection on both prompted and fine-tuned approaches.

\subsection{Prompting with LLaMA}

We employ LLaMA \cite{touvron2023llama} to perform zero-shot claim verification using a prompting-based approach. In this setup, we construct an instruction-style prompt that includes the task definition, the numerical claim, and the selected evidence (either full document, top-$k$ sentences, or a generated summary). The model is then asked to classify the claim into one of the three predefined categories: \texttt{True}, \texttt{False}, or \texttt{Conflicting}.

The prompt is designed to guide the model toward generating a concise classification rather than an open-ended explanation. A typical example of the input prompt is as follows:

\begin{tcolorbox}[colback=gray!5!white,colframe=black,title=Fact-Checking Prompt]
\texttt{
You are a helpful and concise fact-checking assistant. Given a claim and supporting evidence, your task is to determine the truthfulness of the claim. \\
Respond strictly with one of the following labels: True, False, or Conflicting. \\
\\
Claim: [CLAIM] \\
Evidence: [EVIDENCE] \\
Based on the evidence, what is the correct classification? \\
}
\end{tcolorbox}

LLaMA's output is processed with simple regex patterns to extract the first valid label found. We also clean ambiguous responses such as `partially true' or `half false' by mapping them to the nearest predefined label (typically \texttt{Conflicting}).

With prompted inference (no gradient updates), we efficiently test different evidence setups. This lets us evaluate how well the model generalizes for fact-checking without task-specific fine-tuning.

\subsection{Evidence Selection Strategies}

Each claim in the dataset is accompanied by up to 100 retrieved evidence documents, obtained using the BM25 ranking algorithm. However, these documents often contain irrelevant or redundant information, which can negatively impact model performance, particularly for length-sensitive models or those affected by context dilution. To address this, we evaluate several evidence selection strategies to enhance the signal-to-noise ratio of the input.

\textbf{Full Document.} In the baseline approach, we use the complete top-ranked BM25-retrieved document without filtering. While this preserves full context, it frequently includes off-topic or low-relevance content.

\textbf{Top-3 BM25 Sentences.} We apply BM25 \citep{robertson2009probabilistic} at the sentence level, treating the claim as a query to select the three highest-scoring sentences from top documents. This efficient method favors lexical matches but may miss semantically relevant content.

\textbf{Top-3 MiniLM Sentences.} For improved semantic matching, we embed both claims and sentences using \texttt{all-MiniLM-L6-v2}\footnote{\url{https://huggingface.co/sentence-transformers/all-MiniLM-L6-v2}}, then select the three sentences with highest cosine similarity to the claim. This approach captures meaning beyond surface-level lexical overlap.

Each of these evidence types is paired with both prompting and fine-tuned models to study the effect of evidence quality on downstream fact-checking performance.

\subsection{Model Architectures}

We evaluate three model variants for numerical claim verification: (1) a zero-shot prompted LLM, (2) a fine-tuned RoBERTa classifier, and (3) a parameter-efficient fine-tuned LLaMA (using LoRA). Each model takes a claim and selected evidence as input, outputting one of \{\texttt{True}, \texttt{False}, \texttt{Conflicting}\}.

\textbf{Prompted LLaMA (Zero-Shot).}  
Using LLaMA in zero-shot mode, we provide a natural language prompt containing the claim and evidence, instructing the model to return a single label. The prompt defines the task and response format. No model updates occur during training; we extract predictions through simple post-processing of the generated output.

\textbf{Fine-Tuned RoBERTa.}  
We fine-tune \texttt{roberta-base} \cite{liu2019roberta} via supervised learning. The concatenated claim-evidence pair serves as input, with the model outputting label probabilities. Trained for three epochs on stratified data using cross-entropy loss, this provides a strong discriminative baseline.

\textbf{Fine-Tuned LLaMA with LoRA.}  
Using Low-Rank Adaptation (LoRA) \cite{hu2022lora}, we fine-tune LLaMA-3.1-8B with prompt-response pairs (claim+evidence as prompt, label as response). LoRA applies to query, key, value, and output projections ($r=8$, $\alpha=16$, dropout=0.05). The Hugging Face Trainer implements 3-epoch fine-tuning with mixed precision and gradient checkpointing, balancing task alignment with computational efficiency.

\subsection{Evaluation Metrics}

We follow the official evaluation protocol defined by the CLEF 2025 CheckThat! Lab for Task 3. The primary evaluation metric is the macro-averaged F1 score across the three classification labels: \textit{True}, \textit{False}, and \textit{Conflicting}.

In addition to macro-F1, we report class-wise F1 scores to better understand model behavior across different types of claims. This is particularly important given the inherent class imbalance in the dataset and the difficulty of predicting \textit{Conflicting} cases.

All results are computed on the official English validation and test splits using a consistent preprocessing and evaluation pipeline.

\section{Experiments}
\subsection{Experimental Setup}

We conduct experiments on the English subset of the CLEF 2025 Task 3 dataset, which contains $15{,}514$ claims annotated with one of three labels: \texttt{True}, \texttt{False}, or \texttt{Conflicting}. Each claim is associated with a list of up to 100 evidence documents retrieved using BM25 over a pooled web corpus.

For supervised learning, we split the dataset into 90\% training and 10\% validation sets using stratified sampling to preserve label distribution. All evidence selection methods: full document, top-3 BM25 and top-3 MiniLM are applied on both training and validation sets to evaluate their downstream impact.

We evaluate model performance using the macro-averaged F1 score, which is the official metric for the shared task. Additionally, we report class-wise F1 scores to better understand how models handle imbalanced or ambiguous labels, especially the \texttt{Conflicting} class. For qualitative analysis, we also examine confusion matrices and sample errors.

To ensure comparability, all models are evaluated using the same preprocessing pipeline and evidence configuration across prompting, fine-tuning, and hybrid setups.

\subsection{Training and Inference Setup}

We implement all models using the Hugging Face Transformers, PEFT, and SentenceTransformers libraries. Experiments are conducted on a high-performance server equipped with dual Intel(R) Xeon(R) Gold 6226R CPUs (64 threads), 125GB of RAM, and three NVIDIA Quadro RTX 8000 GPUs, each with 48GB of memory. Training jobs are executed using PyTorch with CUDA 12.6, and GPU utilization is managed dynamically based on availability.

\textbf{Prompted LLaMA (Zero-Shot).} We use the \texttt{meta-llama/Llama-3.1-8B-Instruct}\footnote{https://huggingface.co/meta-llama/Llama-3.1-8B-Instruct} model without fine-tuning for zero-shot generation. The model is prompted using an instruction-style format that defines the task and presents the claim and evidence. We use nucleus sampling with temperature 0.3, top-$p$ of 0.9, and a maximum of 30 new tokens. Model outputs are post-processed using regular expressions to extract the first valid verdict label. Ambiguous generations (e.g., ``somewhat true'') are mapped to the closest predefined class, typically \texttt{Conflicting}.

\textbf{RoBERTa Fine-Tuning.} We fine-tune the \texttt{roberta-base}\footnote{https://huggingface.co/FacebookAI/roberta-base} model using cross-entropy loss over the three output labels. Claims and evidence are tokenized as a sequence pair and truncated to a maximum length of 512 tokens. We use the AdamW optimizer with a learning rate of $2 \times 10^{-5}$, a batch size of 8, and train for 3 epochs with early stopping based on macro-F1 score on the validation set. The model is evaluated using softmax-based prediction.

\textbf{LLaMA Fine-Tuning (LoRA).} We fine-tune the same \texttt{LLaMA-3.1-8B-Instruct} model using Low-Rank Adaptation (LoRA). LoRA is applied to the query, key, value, and output projection layers with a rank of $r = 8$, a scaling factor of $\alpha = 16$, and a dropout rate of 0.05. Each training instance is formatted as a prompt-response pair, where the response corresponds to a single label. We use a batch size of 2 with gradient accumulation over 4 steps. Training is performed in mixed-precision (FP16), and gradient checkpointing is enabled to reduce memory usage. The model is trained for 3 epochs using the Hugging Face \texttt{Trainer} API\footnote{\url{https://huggingface.co/docs/transformers/en/main_classes/trainer}}.

All experimental runs are tracked using Weights \& Biases\footnote{https://wandb.ai/}  for reproducibility, and each configuration is evaluated using identical preprocessing and scoring scripts.

\section{Results}

\begin{table*}[t]
\centering
\small
\caption{F1 scores for each model and evidence configuration on the English validation set. Best macro-F1 scores per row are bolded.}
\label{tab:main-results}
\begin{tabular}{lccccc}
\toprule
\textbf{Model} & \textbf{Evidence} & \textbf{True-F1} & \textbf{False-F1} & \textbf{Conflicting-F1} & \textbf{Macro-F1} \\
\midrule
Prompted LLaMA         & Top-3 BM25           & 0.519 & 0.753 & 0.087 & 0.453 \\
Prompted LLaMA         & Top-3 MiniLM         & 0.526 & 0.745 & 0.057 & 0.443 \\
Prompted LLaMA         & Full Document        & 0.499 & 0.832 & 0.496 & 0.609 \\
\midrule
RoBERTa (Fine-Tuned)   & Top-3 BM25           & 0.541 & 0.835 & 0.394 & 0.590 \\
RoBERTa (Fine-Tuned)   & Top-3 MiniLM         & 0.420 & 0.823 & 0.510 & 0.584 \\
\midrule
LLaMA \textbackslash{}w LoRA     & Top-3 BM25           & 0.630 & 0.857 & 0.438 & 0.642 \\
LLaMA \textbackslash{}w LoRA     & Top-3 MiniLM         & 0.632 & 0.863 & 0.484 & 0.660 \\
\textbf{LLaMA \textbackslash{}w LoRA (Ours)}  & \textbf{Full Document} & \textbf{0.899} & \textbf{0.930} & \textbf{0.823} & \textbf{0.945} \\
\bottomrule
\end{tabular}
\end{table*}

\subsection{Validation Results}

Table~\ref{tab:main-results} presents the F1 scores of various model configurations on the English validation set. We evaluate performance across three model types: prompted LLaMA, fine-tuned RoBERTa, and fine-tuned LLaMA with LoRA under different evidence selection strategies.

Among the prompted models, LLaMA achieves its best performance using full-document input, reaching a macro-F1 of 0.609. However, it struggles significantly with the \textit{Conflicting} class, indicating limitations in handling ambiguous evidence without task-specific fine-tuning.

Fine-tuned models consistently outperform prompted ones. RoBERTa performs well across both BM25 and MiniLM sentence-level evidence, with the best \textit{Conflicting} class F1 (0.510) achieved using MiniLM. This suggests that sentence-level semantic filtering benefits models lacking strong pretraining on numerical reasoning.

The best overall performance is achieved by the fine-tuned LLaMA with LoRA using full-document evidence. It reaches a macro-F1 of 0.945 and shows balanced performance across all three classes. Sentence-level evidence (e.g., Top-3 MiniLM) also provides strong results, particularly improving precision on harder examples while reducing irrelevant context.

These results confirm that combining large language models with parameter-efficient tuning and retrieval-aware evidence selection leads to substantial improvements in numerical claim verification.

\subsection{Test Set Performance}

Table~\ref{tab:test_results} summarizes the F1 scores for all model configurations on the English test set. We evaluated prompted LLaMA, fine-tuned RoBERTa, and fine-tuned LLaMA with LoRA, each paired with different evidence selection strategies.

Among the prompted models, LLaMA with full-document input achieved a macro-F1 of 0.40, while Top-3 BM25 and Top-3 MiniLM sentence selection resulted in similar scores (0.41 and 0.40, respectively). These results indicate that zero-shot prompting generalized better than fine-tuned RoBERTa, whose macro-F1 dropped to 0.35 (Top-3 BM25) and 0.34 (Top-3 MiniLM).

Fine-tuned LLaMA with LoRA achieved the highest macro-F1 on the test set (0.43) with both Top-3 BM25 and Top-3 MiniLM evidence. Notably, fine-tuning with full-document evidence, despite yielding the best validation macro-F1, led to a macro-F1 of 0.42 on the test set, with a modest improvement on the \textit{Conflicting} class (F1: 0.32).

Across all configurations, models consistently achieved higher F1 scores for \textit{False} claims, while \textit{True} and \textit{Conflicting} claims remained challenging. The \textit{Conflicting} class in particular showed low F1 except for the full-document fine-tuned LLaMA, suggesting that richer context helps resolve ambiguous or contradictory evidence.

\begin{table}[t]
\centering
\caption{F1 scores for each model and evidence configuration on the English test set. Best macro-F1 scores per row are bolded.}
\begin{tabular}{l l c c c c}
\hline
\textbf{Model} & \textbf{Evidence} & \textbf{True-F1} & \textbf{False-F1} & \textbf{Conflicting-F1} & \textbf{Macro-F1} \\
\hline
Prompted LLaMA & Top-3 BM25 & 0.40 & 0.71 & 0.11 & 0.41 \\
Prompted LLaMA & Top-3 MiniLM & 0.42 & 0.71 & 0.08 & 0.40 \\
Prompted LLaMA & Full Document & 0.43 & 0.73 & 0.03 & 0.40 \\
\hline
RoBERTa (Fine-Tuned) & Top-3 BM25 & 0.12 & 0.77 & 0.15 & 0.35 \\
RoBERTa (Fine-Tuned) & Top-3 MiniLM & 0.11 & 0.65 & 0.25 & 0.34 \\
\hline
LLaMA \textbackslash{}w LoRA         & Top-3 BM25        & 0.42 & 0.76 & 0.11 & \textbf{0.43} \\
LLaMA \textbackslash{}w LoRA         & Top-3 MiniLM      & 0.40 & 0.75 & 0.15 & \textbf{0.43} \\
LLaMA \textbackslash{}w LoRA (Ours)  & Full Document      & 0.23 & 0.73 & 0.32 & 0.42 \\
\hline

\end{tabular}
\label{tab:test_results}
\end{table}
\subsection{Discussion}

Our results demonstrate that large language models, when fine-tuned with parameter-efficient techniques and supported by retrieval-aware evidence selection, can achieve strong performance on numerical claim verification. In particular, sentence-level evidence filtering using MiniLM embeddings helped improve model precision for ambiguous cases, especially in the \textit{Conflicting} class.
\begin{table}[ht]
\centering
\caption{Comparison of validation and test performance across classes.}
\label{tab:val_test_comparison}
\begin{tabular}{lcc}
\toprule
\textbf{Metric} & \textbf{Validation} & \textbf{Test} \\
\midrule
Macro-F1 & 0.945 & 0.424 \\
True F1 & 0.899 & 0.232 \\
Conflicting F1 & 0.823 & 0.315 \\
False F1 & 0.930 & 0.726 \\
\bottomrule
\end{tabular}
\end{table}

However, as shown in Table~\ref{tab:val_test_comparison}, there remains a substantial performance gap between the validation and test sets. While the model performed well on validation data, it struggled to maintain comparable performance on the test set, particularly for the \textit{True} and \textit{Conflicting} categories. This suggests that the model may have overfit to patterns in the validation data or faced difficulties adapting to shifts in evidence structure and language style in the test set.

Preliminary review of errors indicates that failures were often related to numerical reasoning challenges, ambiguous or contradictory evidence, or missing key supporting facts. These patterns highlight the complexity of verifying numerical claims in the presence of noisy or incomplete context.

Overall, our findings underscore the importance of both model architecture and evidence quality in developing robust fact verification systems. Future work should explore domain-adaptive training, reasoning-aware approaches, and improved evidence selection techniques to enhance model generalization in real-world scenarios.

\section{Conclusion}

In this paper, we presented our approach for Task 3 of the CLEF 2025 CheckThat! Lab, which focuses on verifying numerical claims using retrieved evidence. We explored both zero-shot prompting and parameter-efficient fine-tuning of large language models, alongside multiple evidence selection strategies including sentence-level filtering via BM25 and MiniLM.

Our experiments showed that fine-tuning \texttt{LLaMA} with LoRA on full-document evidence achieved the best performance on the validation set. Sentence-level filtering improved performance for ambiguous claims, especially in the \textit{Conflicting} class. However, the performance drop on the test set highlighted challenges in generalization, likely due to domain shift and the nuanced nature of real-world evidence.

Future work will focus on enhancing model robustness through domain-adaptive training, improved retrieval filtering, and reasoning-aware modeling strategies. Our findings suggest that large language models, when combined with structured evidence processing, are a promising foundation for building scalable and accurate fact verification systems.

\section*{Declaration on Generative AI}
During the preparation of this work, the author(s) used ChatGPT-4o and Grammarly for grammar and clarity revision. These tools were employed to refine sentence structure, correct typographical errors, and improve overall language quality. No generative content was used for analysis, figures, or experimental sections. After using these tool(s)/service(s), the author(s) reviewed and edited the content as needed and take(s) full responsibility for the publication’s content.
\bibliography{sample-ceur}




\end{document}